\documentclass[11pt,a4paper]{article}
\usepackage[hyperref,final]{naacl2021}
\usepackage{times}
\usepackage{latexsym}
\usepackage{booktabs} 
\usepackage{multirow}
\usepackage{amsmath}
\usepackage{amssymb}
\usepackage{graphicx}

\usepackage[utf8]{inputenc} 
\usepackage[T1]{fontenc}    

\usepackage{url}            
\usepackage{booktabs}       
\usepackage{amsfonts}       
\usepackage{nicefrac}       
\usepackage{microtype}      

\usepackage{latexsym}
\usepackage{booktabs} 
\usepackage{multirow}
\usepackage{amsmath}
\usepackage{amssymb}
\usepackage{graphicx}
\usepackage{float}

\usepackage{natbib}

\usepackage{cleveref}
\crefformat{section}{\S#2#1#3}
\crefformat{subsection}{\S#2#1#3}
\crefformat{subsubsection}{\S#2#1#3}
\crefrangeformat{section}{\S\S#3#1#4 to~#5#2#6}
\crefmultiformat{section}{\S\S#2#1#3}{ and~#2#1#3}{, #2#1#3}{ and~#2#1#3}
\usepackage{refstyle}
\Crefformat{figure}{#2Fig.~#1#3}
\Crefmultiformat{figure}{Figs.~#2#1#3}{ and~#2#1#3}{, #2#1#3}{ and~#2#1#3}
\Crefformat{table}{#2Tab.~#1#3}
\Crefmultiformat{table}{Tabs.~#2#1#3}{ and~#2#1#3}{, #2#1#3}{ and~#2#1#3}
\Crefformat{equation}{#2Eq.~(#1#3)}

\Crefformat{appendix}{#2App.~\S#1#3}

\usepackage{colortbl}
\newcommand{\CC}[1]{\cellcolor{blue!#1}}
\newcommand{\CCR}[1]{\cellcolor{red!#1}}
\newcommand{\CCG}[1]{\cellcolor{green!#1}}

\usepackage{arydshln}
\usepackage{subcaption}

\usepackage{microtype}

\usepackage{pifont}
\newcommand{\cmark}{\ding{51}}%
\newcommand{\xmark}{\ding{55}}%

\newcommand{\stitle}[1]{\vspace{1.8ex} \noindent{\bf #1}}

\title{Self-Alignment Pretraining for Biomedical Entity Representations}

\author{Fangyu Liu$^{\clubsuit}$, Ehsan Shareghi$^{\diamondsuit,\clubsuit}$, Zaiqiao Meng$^{\clubsuit}$, Marco Basaldella$^{\heartsuit}$\thanks{$\ \ $Work conducted prior to joining Amazon.} , Nigel Collier$^\clubsuit$ \\
  $^\clubsuit$Language Technology Lab, TAL, University of Cambridge \\
  $^\diamondsuit$Department of Data Science \& AI, Monash University   $\ $ $^\heartsuit$Amazon Alexa \\
  $^\clubsuit$\texttt{\{fl399, zm324, nhc30\}@cam.ac.uk}\\ $^\diamondsuit$\texttt{ehsan.shareghi@monash.edu}
  $^\heartsuit$\texttt{mbbasald@amazon.co.uk}
}

\date{}

\begin{document}
\maketitle


\begin{abstract}

Despite the widespread success of self-supervised learning via masked language models (MLM), 
accurately capturing fine-grained
semantic relationships in the biomedical domain remains a challenge. This is of paramount importance for entity-level tasks such as entity linking where the ability to model entity relations (especially synonymy) is pivotal. 
{To address this challenge, we propose \textsc{SapBert}, a pretraining scheme that self-aligns the representation space of biomedical entities. We design a scalable metric learning framework that can leverage UMLS, a massive collection of biomedical ontologies with 4M+ concepts.} 
In contrast with previous pipeline-based hybrid systems, \textsc{SapBert} offers an elegant one-model-for-all solution to the problem of medical entity linking (MEL), achieving a new state-of-the-art (SOTA) on six MEL benchmarking datasets. In the scientific domain, we achieve SOTA even without task-specific supervision. With substantial improvement over various domain-specific pretrained MLMs such as \textsc{BioBert}, \textsc{SciBert} and \textsc{PubMedBert}, our pretraining scheme proves to be both effective and robust.\footnote{For code and pretrained models, please visit: \url{https://github.com/cambridgeltl/sapbert}.}

\end{abstract}
\section{Introduction}\label{sec:intro}
\begin{figure}
    \includegraphics[width=\linewidth]{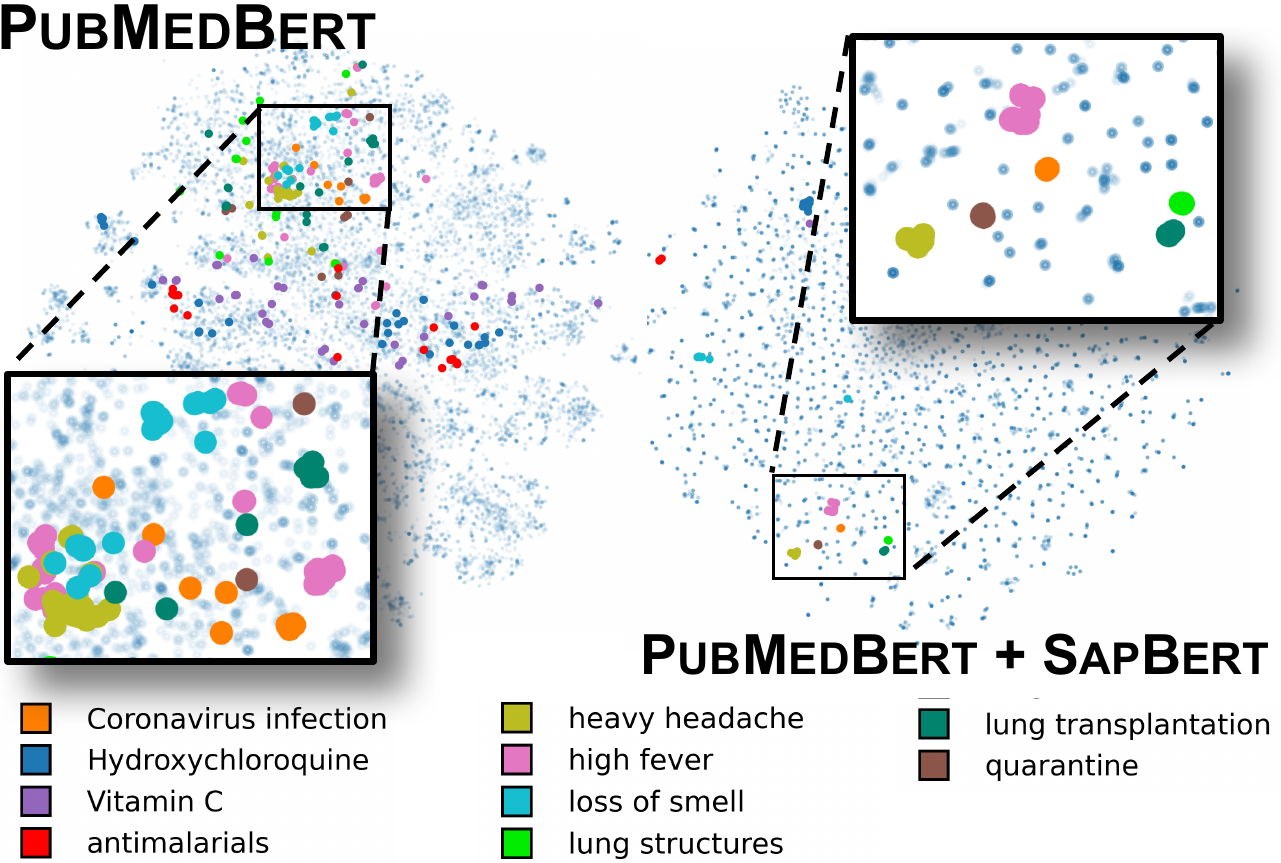}
    \caption{The t-SNE \citep{maaten2008visualizing} visualisation of UMLS entities under \textsc{PubMedBert} (\textsc{Bert} pretrained on PubMed papers) \& \textsc{PubMedBert}+\textsc{SapBert} (\textsc{PubMedBert} further pretrained on UMLS synonyms). The biomedical names of different concepts are hard to separate in the heterogeneous embedding space (left). After the self-alignment pretraining, the same concept's entity names are drawn closer to form compact clusters (right).}
    \label{fig:front}
\end{figure}

Biomedical entity\footnote{In this work, \emph{biomedical entity} refers to the surface forms of biomedical concepts, which can be a single word (e.g. \emph{fever}), a compound (e.g. \emph{sars-cov-2}) or a short phrase (e.g. \emph{abnormal retinal vascular development}).} representation is the foundation for a plethora of text mining systems in the medical domain, facilitating applications such as literature search \citep{lee2016best}, clinical decision making \citep{roberts2015overview} and relational knowledge discovery (e.g. chemical-disease, drug-drug and protein-protein relations, \citealt{wang2018comparison}). 
The heterogeneous naming of biomedical concepts poses a major challenge to representation learning. For instance, the medication \emph{Hydroxychloroquine} 
is often referred to as \emph{Oxichlorochine} (alternative name), \emph{HCQ} (in social media) and \emph{Plaquenil} (brand name). 

MEL addresses 
this problem by framing it as a task of mapping entity mentions to unified concepts in a medical knowledge graph.\footnote{Note that we consider only the biomedical entities themselves and not their contexts, also known as medical concept normalisation/disambiguation in the BioNLP community.} The main bottleneck of MEL is the quality of the entity representations ~\cite{cometa}. Prior works in this domain have adopted very sophisticated text pre-processing heuristics \cite{d2015sieve,kim2019neural,ji2020bert, sung2020biomedical} which can hardly cover all the variations of biomedical names. 
In parallel, self-supervised learning has shown tremendous success
in NLP via leveraging the masked language modelling (MLM) objective to learn semantics from distributional representations \citep{devlin2019bert,liu2019roberta}.
{Domain-specific pretraining on biomedical corpora (e.g. \textsc{BioBert}, \citealt{lee2020biobert} and \textsc{BioMegatron}, \citealt{shin-etal-2020-biomegatron}) have made much progress in biomedical text mining tasks.} Nonetheless, representing medical entities with the existing SOTA pretrained MLMs (e.g. \textsc{PubMedBert}, \citealt{pubmedbert}) 
as suggested in \Cref{fig:front} (left) does not lead to a well-separated representation space. 

To address the aforementioned issue, we propose to pretrain a Transformer-based language model on the biomedical knowledge graph of UMLS \cite{bodenreider2004unified}, 
the largest interlingua of biomedical ontologies. UMLS contains a comprehensive collection of biomedical synonyms in various forms (UMLS 2020AA has 4M+ concepts and 10M+ synonyms which stem from over 150 controlled vocabularies including MeSH, SNOMED CT, RxNorm, Gene Ontology and OMIM).\footnote{{\scriptsize \url{https://www.nlm.nih.gov/research/umls/knowledge_sources/metathesaurus/release/statistics.html}}} We design a self-alignment objective that clusters synonyms of the same concept. To cope with the immense size of UMLS, we sample hard training pairs from the knowledge base and use a scalable metric learning loss. We name our model as \textbf{S}elf-\textbf{a}ligning \textbf{p}re-trained \textbf{\textsc{Bert}} (\textsc{SapBert}).


 
 
 Being both simple and powerful, \textsc{SapBert} obtains new SOTA performances across all six MEL benchmark datasets. 
In contrast with the current systems which adopt complex pipelines and hybrid components \citep{xu-etal-2020-generate, ji2020bert, sung2020biomedical}, \textsc{SapBert} applies a much simpler training procedure without requiring any pre- or post-processing steps. At test time, a simple nearest neighbour's search is sufficient for making a prediction. 
When compared with other domain-specific pretrained language models (e.g. \textsc{BioBert}
and \textsc{SciBert}), 
\textsc{SapBert} also brings substantial improvement by {up to 20\%} on accuracy across all tasks. The effectiveness of the pretraining in \textsc{SapBert} is especially highlighted in the scientific language domain where \textsc{SapBert} outperforms previous SOTA even without fine-tuning on any MEL datasets. 
We also provide insights on pretraining's impact across domains and explore pretraining with fewer model parameters by using a recently introduced \textsc{Adapter} module in our training scheme. 


\section{Method: Self-Alignment Pretraining}\label{sec:method}
{We design a metric learning framework that learns to self-align synonymous biomedical entities. The framework can be used as both pretraining on UMLS, and fine-tuning on task-specific datasets. We use an existing \textsc{Bert} model as our starting point. In the following, we introduce the key components of {our} framework. }

\stitle{Formal Definition.} 
Let $(x, y)\in \mathcal{X} \times \mathcal{Y}$ denote a tuple of a name and its categorical label.
For the self-alignment pretraining step, $\mathcal{X} \times \mathcal{Y}$ is the set of all (name, CUI\footnote{In UMLS, CUI is the \textbf{C}oncept \textbf{U}nique \textbf{I}dentifier.}) pairs in UMLS, e.g. (\emph{Remdesivir},  C4726677); while for the fine-tuning step, it is formed as an entity mention and its corresponding mapping from the ontology, e.g. (\emph{scratchy throat}, 102618009).
Given any pair of tuples $(x_i, y_i),(x_j, y_j) \in \mathcal{X}\times \mathcal{Y}$, the goal of the self-alignment is to learn a function $f(\cdot;\theta): \mathcal{X} \rightarrow \mathbb{R}^d$ 
parameterised by $\theta$. Then, the similarity 
$\langle f(x_i),f(x_j) \rangle$ (in this work we use cosine similarity) can be used to estimate the resemblance of $x_i$ and $x_j$ (i.e., high if $x_i,x_j$ are synonyms and low otherwise). We model $f$ by a \textsc{Bert} model with its output \texttt{[CLS]} token regarded as the representation of the input.\footnote{We tried multiple strategies including first-token, mean-pooling, \texttt{[CLS]} and also \textsc{nospec} (recommended by \citealt{vulic-etal-2020-probing}) but found no consistent best strategy (optimal strategy varies on different \textsc{*Bert}s).} During the learning, a sampling procedure selects the informative pairs of training samples and uses them in the pairwise metric learning loss function (introduced shortly).

\begin{figure}[t!]
\centering
    \includegraphics[width=0.45\linewidth]{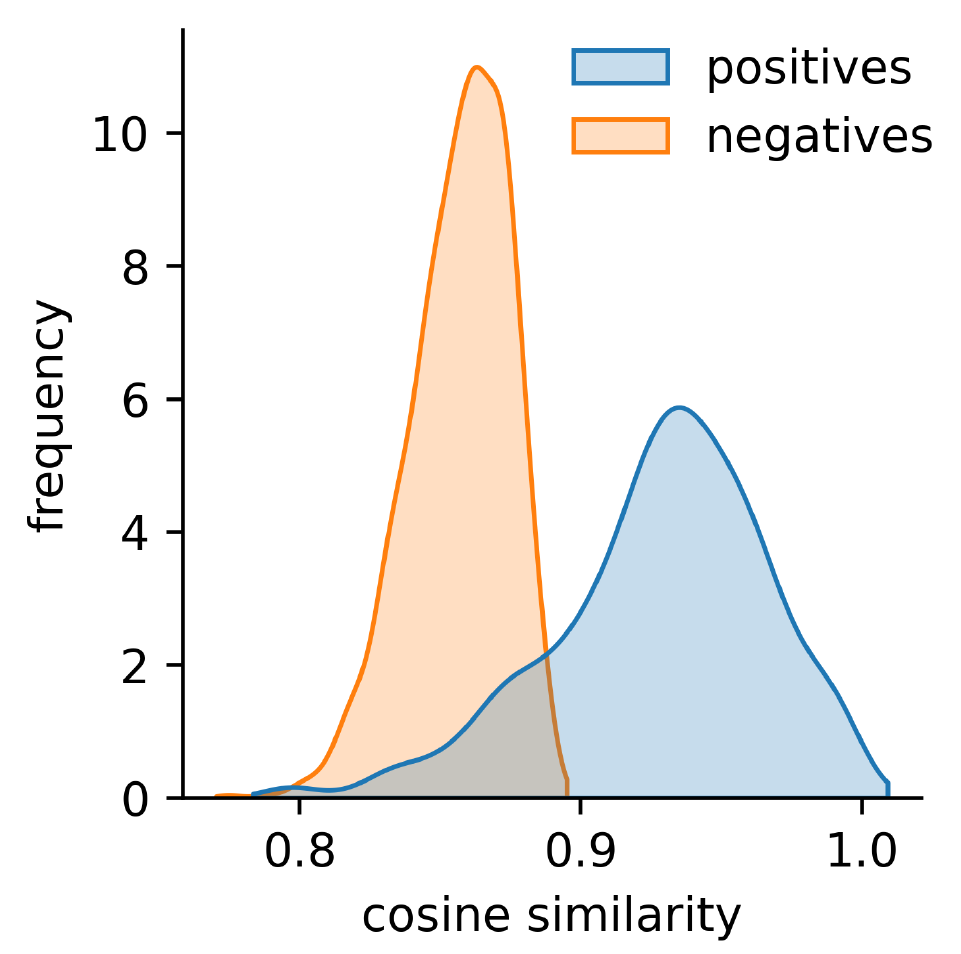}
    \includegraphics[width=0.45\linewidth]{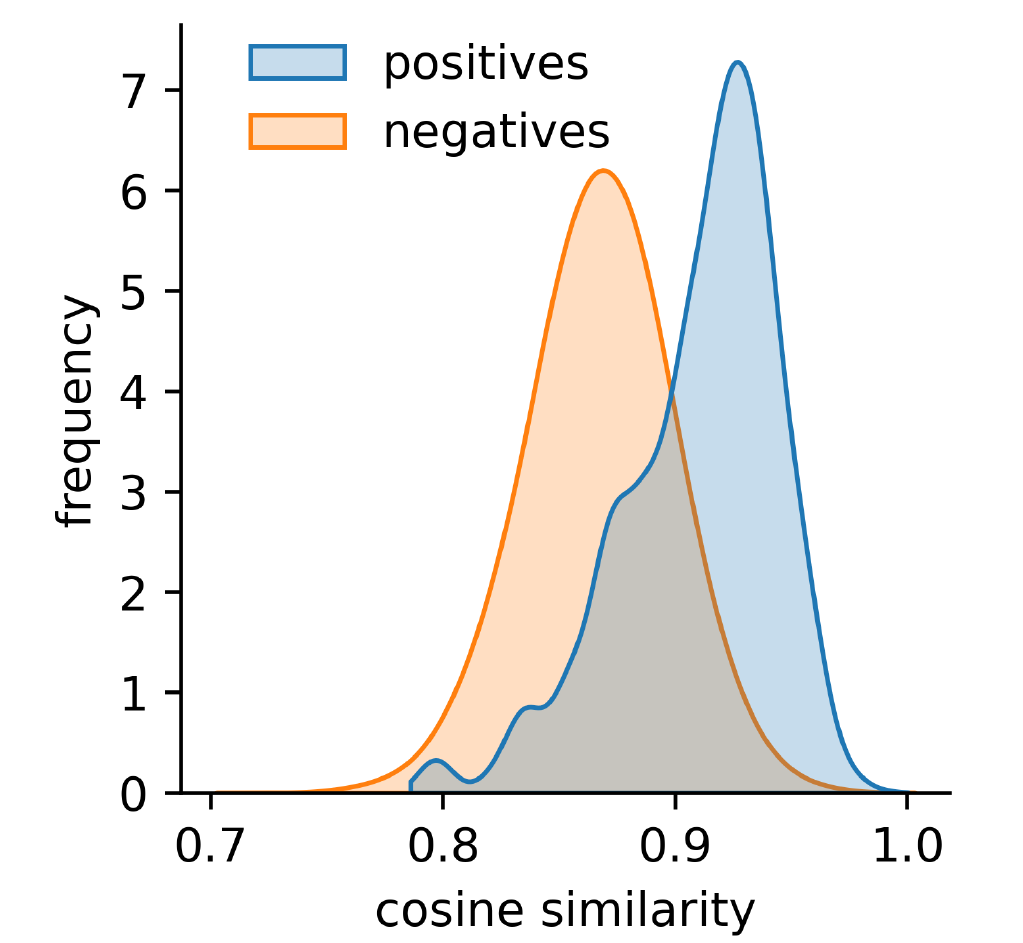}
    \caption{
    The distribution of similarity scores for all sampled \textsc{PubMedBert} representations in a mini-batch. The left graph shows the distribution of \textbf{+} and \textbf{-} pairs which are easy and already well-separated. The right graph illustrates larger overlap between the two groups generated by the online mining step, making them harder and more informative for learning.}
  \label{fig:mining}
\end{figure}

\stitle{Online Hard Pairs Mining.} We use an online hard triplet mining condition to find the most informative training examples (i.e. hard positive/negative pairs) within a mini-batch for efficient training, \Cref{fig:mining}. For biomedical entities, this step can be particularly useful as most examples can be easily classified while a small set of very hard ones cause the most challenge to representation learning.\footnote{Most of \emph{Hydroxychloroquine}'s variants are easy: \emph{Hydroxychlorochin}, \emph{Hydroxychloroquine (substance)}, \emph{Hidroxicloroquina}, but a few can be very hard: \emph{Plaquenil} and \emph{HCQ}.} 
We start from constructing all possible triplets for all names within the mini-batch where each triplet is in the form of $(x_a, x_p, x_n)$. Here $x_a$ is called \emph{anchor}, an arbitrary name in the mini-batch; $x_p$ a positive match of $x_a$ (i.e. $y_a=y_p$) and $x_n$ a negative match of $x_a$ (i.e. $y_a \neq y_n$). Among the constructed triplets, we select out all triplets that violate the following condition:
\begin{equation}
\setlength{\abovedisplayskip}{7pt}
\setlength{\belowdisplayskip}{7pt}
    \|f(x_a)-f(x_p)\|_2 <  \|f(x_a)-f(x_n)\|_2 + \lambda,
    \label{eq:mining}
\end{equation}
where $\lambda$ is a pre-set margin. In other words, we only consider triplets with the negative sample closer to the positive sample by a margin of $\lambda$. 
These are the hard triplets as their original representations were very far from correct. Every hard triplet contributes one hard positive pair $(x_a,x_p)$ and one hard negative pair $(x_a,x_n)$. We collect all such positive \& negative pairs and denote them as $\mathcal{P},\mathcal{N}$. A similar but not identical triplet mining condition was used by \citet{schroff2015facenet} for face recognition to select hard negative samples. Switching-off this mining process, causes a drastic performance drop (see \Cref{Table:w_or_wo_mining}).

\stitle{Loss Function.} We compute the pairwise cosine similarity of all the \textsc{Bert}-produced name representations and obtain a similarity matrix $\mathbf{S}\in\mathbb{R}^{|\mathcal{X}_b|\times |\mathcal{X}_b|}$ where each entry $\mathbf{S}_{ij}$ corresponds to the cosine similarity between the $i$-th and $j$-th names in the mini-batch $b$. We adapted the Multi-Similarity loss~(MS loss, \citealt{wang2019multi}), a SOTA metric learning objective on visual recognition, for learning from the positive and negative pairs:
\begin{equation}
\setlength{\abovedisplayskip}{6pt}
\setlength{\belowdisplayskip}{6pt}
    \begin{split}
\mathcal{L} = \frac{1}{|\mathcal{X}_b|}\sum_{i=1}^{|\mathcal{X}_b|} \Bigg(\frac{1}{\alpha}\log\Big(1+\sum_{n\in \mathcal{N}_i } e^{\alpha (\mathbf{S}_{in} -\epsilon )}\Big) \\
 + \frac{1}{\beta} \log \Big(1+\sum_{p \in \mathcal{P}_{i}} e ^{ - \beta (\mathbf{S}_{ip} -\epsilon)} \Big)\Bigg) ,
\label{eq:loss}
\end{split}
\end{equation}
where $\alpha,\beta$ are temperature scales; $\epsilon$ is an offset applied on the similarity matrix; $\mathcal{P}_i,\mathcal{N}_i$ are indices of positive and negative samples of the \emph{anchor} $i$.\footnote{We explored several loss functions such as InfoNCE \citep{oord2018representation}, NCA loss~\citep{goldberger2005neighbourhood}, simple cosine loss~\citep{phan2019robust}, max-margin triplet loss~\citep{cometa} but found our choice is empirically better. See \Cref{sec:compare_loss} for comparison.}

While the first term in Eq.~\ref{eq:loss} pushes negative pairs away from each other, the second term pulls positive pairs together. This dynamic allows for a  re-calibration of the alignment space using the semantic biases of synonymy relations.
%
%
The MS loss leverages similarities among and between positive and negative pairs to re-weight the importance of the samples.
The most informative pairs will receive more gradient signals during training and thus can better use the information stored in data. 


\section{Experiments and Discussions}\label{sec:experiment}

\begin{table*}[t]
\setlength{\tabcolsep}{3.2pt}
\centering
\scriptsize
\centering
\vspace{-1.6em}
\begin{tabular}{lccccccccccccccccccc}
\toprule
& & \multicolumn{11}{c}{\small scientific language}  & \multicolumn{5}{c}{\small social media language}\\
\cmidrule[1.0pt]{2-12} \cmidrule[1.0pt]{14-18}
 \multirow{2}{*}{\small model} & \multicolumn{2}{c}{NCBI} &  &\multicolumn{2}{c}{BC5CDR-d} &  &\multicolumn{2}{c}{BC5CDR-c} & & \multicolumn{2}{c}{MedMentions} & & \multicolumn{2}{c}{AskAPatient} & & \multicolumn{2}{c}{COMETA}   \\
\cmidrule[1.5pt]{2-3}\cmidrule[1.5pt]{5-6}\cmidrule[1.5pt]{8-9}\cmidrule[1.5pt]{11-12}\cmidrule[1.5pt]{14-15}\cmidrule[1.5pt]{17-18}
 & $@1$ & $@5$ & & $@1$ & $@5$ & & $@1$ & $@5$ & & $@1$ & $@5$  & & $@1$ & $@5$ & & $@1$ & $@5$ \\
\cmidrule[1.0pt]{1-20}
  vanilla \textsc{Bert} \citep{devlin2019bert} & 67.6 & 77.0 & &  81.4 & 89.1 & & 79.8 & 91.2 & &39.6 & 60.2 &  & 38.2 & 43.3   & &40.4 & 47.7 \\
 {\ \ \ \ + \textsc{SapBert}} & \CCG{48}91.6 & \CCG{36.4}95.2 & & \CCG{22.6}92.7 & \CCG{12.6}95.4 & & \CCG{32.6}96.1 & \CCG{13.6}98.0 & & \CCG{25.8}52.5 & \CCG{24.8}72.6& & \CCG{60.4}68.4 & \CCG{88.6}87.6 & & \CCG{38.2}59.5 & \CCG{58.2}76.8 \\\hdashline
 \textsc{BioBert} \citep{lee2020biobert} & 71.3 & 84.1 & & 79.8 & 92.3 & & 74.0 & 90.0 & &24.2 & 38.5 & &41.4 & 51.5  & & 35.9 & 46.1 \\
 {\ \ \ \ + \textsc{SapBert}}  & \CCG{39.4}91.0 & \CCG{21.2}94.7 & & \CCG{27}93.3 & \CCG{24.4}95.5 & & \CCG{45.2}96.6 & \CCG{15.2}97.6 & & \CCG{57.4}53.0 & \CCG{70.4}73.7 & & \CCG{62}72.4 & \CCG{75.2}89.1 & & \CCG{54.8}63.3 & \CCG{61.8}77.0 \\\hdashline
 \textsc{BlueBert} \citep{peng2019transfer}  & 75.7 & 87.2 & & 83.2 & 91.0 & & 87.7 & 94.1 & &41.6 & 61.9 & & 41.5 & 48.5  & & 42.9 & 52.9 \\
 {\ \ \ \ + \textsc{SapBert}}  & \CCG{30.4}90.9 & \CCG{13.6}94.0 & & \CCG{20.4}93.4 & \CCG{10}96.0 & & \CCG{18}96.7 & \CCG{8.2}98.2 & & \CCG{16}49.6 & \CCG{22.4}73.1 & & \CCG{61.8}72.4 & \CCG{81.8}89.4 & & \CCG{46.2}66.0 & \CCG{51.8}78.8 \\\hdashline
 \textsc{ClinicalBert} \citep{clinicalbert} & 72.1 & 84.5 & & 82.7 & 91.6 & & 75.9 & 88.5 & & 43.9 & 54.3 & & 43.1 & 51.8 & & 40.6 & 61.8 \\
 {\ \ \ \ + \textsc{SapBert}}  & \CCG{38}91.1 & \CCG{21.2}95.1 & & \CCG{20.6}93.0 & \CCG{8.2}95.7 & & \CCG{41.4}96.6 & \CCG{18.4}97.7 & & \CCG{15.2}51.5  & \CCG{37.4}73.0 & & \CCG{68}71.1  & \CCG{73.4}88.5 & & \CCG{47.4}64.3 & \CCG{31.0}77.3\\\hdashline
 \textsc{SciBert} \citep{Beltagy2019SciBERT}  &  85.1 & 88.4 & & 89.3 & 92.8 & & 94.2 & 95.5 & & 42.3 & 51.9 & & 48.0 & 54.8 & & 45.8 & 66.8\\
 {\ \ \ \ + \textsc{SapBert}}  & \CCG{13.2}91.7 & \CCG{13.6}95.2 & & \CCG{8}93.3 & \CCG{5.8}95.7 & & \CCG{4.8}96.6 & \CCG{5}98.0 & & \CCG{15.6}50.1 & \CCG{44}73.9 & & \CCG{48.2}72.1  & \CCG{67.8}88.7 & & \CCG{37.4}64.5 & \CCG{21.4}77.5 \\\hdashline
  \textsc{UmlsBert} \citep{umlsbert} & 77.0 & 85.4 && 85.5 & 92.5 && 88.9 & 94.1 && 36.1 & 55.8 && 44.4 & 54.5 && 44.6 & 53.0 \\
 {\ \ \ \ + \textsc{SapBert}} &  \CCG{28.4}91.2 & \CCG{19.6}95.2 && \CCG{14.6}92.8 & \CCG{6}95.5 && \CCG{15.4}96.6 & \CCG{7.2}97.7 && \CCG{32}52.1 & \CCG{34.8}73.2 && \CCG{56.4}72.6 & \CCG{69.6}89.3 && \CCG{37.6}63.4 & \CCG{47.8}76.9\\\hdashline
 \textsc{PubMedBert} \citep{pubmedbert} & 77.8 & 86.9 & & 89.0 & 93.8 & & 93.0 & 94.6 & & 43.9 & 64.7& & 42.5 & 49.6 & & 46.8 & 53.2 \\
 {\ \ \ \ + \textsc{SapBert}} & \CCG{28.4}92.0 & \CCG{17.4}95.6 & & \CCG{9}93.5 & \CCG{4.4}96.0 & & \CCG{7}96.5 & \CCG{7.2}98.2 & & \CCG{13.8}50.8 & \CCG{19.4}74.4 & & \CCG{56}70.5& \CCG{39.3}88.9 & & \CCG{38.2}65.9 & \CCG{49.4}77.9 \\
\cmidrule[1.0pt]{1-18}
\CCR{10}supervised SOTA & \CCR{10}91.1 & \CCR{10}93.9 &\CCR{10} & \CCR{10}93.2 & \CCR{10}96.0 & \CCR{10} & \CCR{10}\underline{96.6} & \CCR{10}97.2 & \CCR{10} & \CCR{10}OOM & \CCR{10}OOM & \CCR{10} & \CCR{10}87.5 \CCR{10}& \CCR{10}- &\CCR{10} & \CCR{10}\underline{79.0} & \CCR{10}- \\ 
\hdashline
\CC{15}\textsc{PubMedBert}& \CC{15}77.8 &\CC{15}86.9 &\CC{15} &\CC{15}89.0 &\CC{15}93.8 &\CC{15} & \CC{15}93.0 &\CC{15}94.6 &\CC{15} & \CC{15}43.9 & \CC{15}64.7 &\CC{15} & \CC{15}42.5 &\CC{15}49.6 &\CC{15} &\CC{15}46.8 &\CC{15}53.2  \\
\CC{15}\ \ \ + \textsc{SapBert} & \CC{15}92.0 & \CC{15}95.6 &\CC{15} & \CC{15}93.5 & \CC{15}96.0 &\CC{15} & \CC{15}96.5 & \CC{15}\underline{98.2} & \CC{15} & \CC{15}\underline{50.8} & \CC{15}74.4 \CC{15} & \CC{15} & \CC{15}70.5 & \CC{15}88.9 &\CC{15} & \CC{15}65.9&\CC{15}77.9\\
\CC{15}\ \ \ + \textsc{SapBert (Adapter$_{13\%}$)} & \CC{15}91.5 & \CC{15}\underline{95.8} &\CC{15} & \CC{15}93.6 & \CC{15}\underline{96.3} &\CC{15} & \CC{15}96.5& \CC{15}	98.0  & \CC{15} & \CC{15}50.7 & \CC{15}\textbf{75.0}$^\dagger$ \CC{15} & \CC{15} & \CC{15}67.5 & \CC{15}87.1 &\CC{15} & \CC{15}64.5&\CC{15}74.9\\
\CC{15}\ \ \ + \textsc{SapBert (Adapter$_{1\%}$)} & \CC{15}90.9& \CC{15}95.4 &\CC{15} & \CC{15}\textbf{93.8}$^\dagger$	& \CC{15}\textbf{96.5}$^\dagger$ &\CC{15} & \CC{15}96.5  & \CC{15}97.9 & \CC{15} & \CC{15}\textbf{52.2}$^\dagger$ & \CC{15}\underline{74.8} & \CC{15} & \CC{15}65.7 & \CC{15}84.0 &\CC{15} & \CC{15}63.5&\CC{15}74.2\\
\CCR{10}\ \ \ + \textsc{SapBert (Fine-tuned)} &\CCR{10}\underline{92.3} &\CCR{10}95.5 &\CCR{10} &\CCR{10}93.2 &\CCR{10}95.4 &\CCR{10} &\CCR{10}96.5 &\CCR{10}97.9 &\CCR{10} &\CCR{10}50.4 &\CCR{10}73.9 &\CCR{10} & \CCR{10}\textbf{89.0}$^\dagger$& \CCR{10}\textbf{96.2}$^\dagger$ &\CCR{10} & \CCR{10}75.1 (\textbf{81.1}$^\dagger$) & \CCR{10}85.5 (\textbf{86.1}$^\dagger$) \\
\hdashline
\CCR{10}\textsc{BioSyn} & \CCR{10}91.1 & \CCR{10}93.9 &\CCR{10} & \CCR{10}93.2 & \CCR{10}96.0 & \CCR{10} & \CCR{10}\underline{96.6} & \CCR{10}97.2 & \CCR{10} & \CCR{10}OOM & \CCR{10}OOM & \CCR{10} & \CCR{10}82.6 \CCR{10}& \CCR{10}87.0 &\CCR{10} & \CCR{10}71.3 & \CCR{10}77.8 \ \\
\CCR{10} \ \ \ + (init. w/) \textsc{SapBert} & \CCR{10}\textbf{92.5}$^\dagger$ & \CCR{10}\textbf{96.2}$^\dagger$ &\CCR{10} & \CCR{10}\underline{93.6} & \CCR{10}96.2 &\CCR{10} & \CCR{10}\textbf{96.8} & \CCR{10}\textbf{98.4}$^\dagger$ &\CCR{10} & \CCR{10}OOM & \CCR{10}OOM &\CCR{10} &\CCR{10}\underline{87.6} &\CCR{10} \underline{95.6} &\CCR{10} &\CCR{10}77.0 &\CCR{10}\underline{84.2} \\
\bottomrule
\end{tabular}
\caption{\textbf{Top}:
Comparison of 7 \textsc{Bert}-based models before and after \textsc{SapBert} pretraining (+ \textsc{SapBert}). All results in this section are from unsupervised learning (not fine-tuned on task data). 
The gradient of \colorbox{green!30}{green} indicates the improvement comparing to the base model (the deeper the more).
\textbf{Bottom}:
\textsc{SapBert} vs. SOTA results. 
\colorbox{blue!15}{Blue} and \colorbox{red!10}{red} denote unsupervised and supervised models. \textbf{Bold} and \underline{underline} denote the best and second best results in the column. ``$^\dagger$'' denotes statistically significant better than supervised SOTA (T-test, $\rho<0.05$).
On COMETA, the results inside the parentheses added the supervised SOTA's dictionary back-off technique~\citep{cometa}. ``-'': not reported in the SOTA paper. ``OOM'': out-of-memory (192GB+).  
}
\label{tab:sap_bert_vs_sota}
\end{table*}

\subsection{Experimental Setups}

\stitle{Data Preparation Details for UMLS Pretraining.}\label{sec:data_prep}
We download the full release of UMLS 2020AA version.\footnote{\scriptsize{\url{https://download.nlm.nih.gov/umls/kss/2020AA/umls-2020AA-full.zip}}} We then extract all English entries from the \texttt{MRCONSO.RFF} raw file and convert all entity names into lowercase (duplicates are removed). 
Besides synonyms defined in \texttt{MRCONSO.RFF}, we also include tradenames of drugs as synonyms (extracted from \texttt{MRREL.RRF}). 
After pre-processing, a list of 9,712,959 (name, CUI) entries is obtained. However, random batching on this list can lead to very few (if not none) positive pairs within a mini-batch. To ensure sufficient positives present in each mini-batch, we generate offline positive pairs in the format of (name$_1$, name$_2$, CUI) where name$_1$ and name$_2$ have the same CUI label. This can be achieved by enumerating all possible combinations of synonym pairs with common CUIs. For balanced training, any concepts with more than 50 positive pairs are randomly trimmed to 50 pairs. In the end we obtain a training list with 11,792,953 pairwise entries.

\stitle{UMLS Pretraining Details.} During training, we use AdamW \citep{loshchilov2018decoupled} with a learning rate of \texttt{2e-5} and weight decay rate of \texttt{1e-2}. Models are trained on the prepared pairwise UMLS data for 1 epoch (approximately 50k iterations) with a batch size of 512 (i.e., 256 pairs per mini-batch). We train with Automatic Mixed Precision (AMP)\footnote{\scriptsize{\url{https://pytorch.org/docs/stable/amp.html}}} provided in PyTorch 1.7.0. This takes approximately 5 hours on our machine (configurations specified in \Cref{sec:appendix_hardware}). For other hyper-parameters used, please view \Cref{sec:hyperparam}.

\stitle{Evaluation Data and Protocol.} 
We experiment on 6 different English MEL datasets: 4 in the scientific domain (NCBI, \citealt{dougan2014ncbi}; BC5CDR-c and BC5CDR-d,~\citealt{li2016biocreative}; MedMentions,~\citealt{mohan2018medmentions}) and 2 in the social media domain (COMETA, \citealt{cometa} and AskAPatient,~\citealt{limsopatham2016normalising}). Descriptions of the datasets and their statistics are provided in \Cref{sec:datasets}. We report Acc$_{@1}$ and Acc$_{@5}$ (denoted as $@1$ and $@5$) for evaluating performance. In all experiments, \textsc{SapBert} denotes further pretraining with our self-alignment method on UMLS. At the test phase,  for all \textsc{SapBert} models we use nearest neighbour search without further fine-tuning on task data (unless stated otherwise). Except for numbers reported in previous papers, all results 
are the average of five runs with different random seeds.

\stitle{Fine-Tuning on Task Data.} The red rows in \Cref{tab:sap_bert_vs_sota} are results of models (further) fine-tuned on the training sets of the six MEL datasets. Similar to pretraining, a positive pair list is generated through traversing the combinations of mention and all ground truth synonyms where mentions are from the training set and ground truth synonyms are from the reference ontology. We use the same optimiser and learning rates but train with a batch size of 256 (to accommodate the memory of 1 GPU). On scientific language datasets, we train for 3 epochs while on AskAPatient and COMETA we train for 15 and 10 epochs respectively. For \textsc{BioSyn} on social media language datasets, we empirically found that 10 epochs work the best. Other configurations are the same as the original \textsc{BioSyn} paper.

\subsection{Main Results and Analysis}
\stitle{\textsc{*Bert + SapBert} (\Cref{tab:sap_bert_vs_sota}, top).} We illustrate the impact of \textsc{SapBert} pretraining over 7 existing \textsc{Bert}-based models (*\textsc{Bert} = \{\textsc{\textsc{BioBert}, PubMedBert}, ...\}).
\textsc{SapBert} obtains consistent improvement over all *\textsc{Bert} models across all datasets, with larger gains (by up to 31.0\% absolute Acc$_{@1}$ increase) observed in the social media domain. While \textsc{SciBert} is the leading model before applying \textsc{SapBert}, \textsc{PubMedBert+SapBert} performs the best afterwards.



\stitle{\textsc{SapBert} vs. SOTA (\Cref{tab:sap_bert_vs_sota}, bottom). } We take \textsc{PubMedBert+SapBert} (w/wo fine-tuning) and compare against various published SOTA results (see \Cref{sec:appendix_full_baselines} for a full listing of 10 baselines) which all require task supervision. For the scientific language domain, the SOTA is \textsc{BioSyn}~\citep{sung2020biomedical}. For the social media domain, the SOTA are \citet{cometa} and \textsc{Gen-Rank}~\citep{xu-etal-2020-generate} on COMETA and AskAPatient respectively. All these SOTA methods combine \textsc{Bert} with heuristic modules such as tf-idf, string matching and information retrieval system (i.e. Apache Lucene) in a multi-stage manner. 

Measured by Acc$_{@1}$, \textsc{SapBert} 
achieves new SOTA with statistical significance on 5 of the 6 datasets and for the dataset (BC5CDR-c) where \textsc{SapBert} is not significantly better, it performs on par with SOTA (96.5 vs. 96.6). 
%
Interestingly, on scientific language datasets, \textsc{SapBert} outperforms SOTA without any task supervision (fine-tuning mostly leads to overfitting and performance drops). On social media language datasets, unsupervised \textsc{SapBert} lags behind supervised SOTA by large margins, highlighting the well-documented complex nature of social media language~\citep{baldwin2013noisy,limsopatham-collier-2015-adapting,limsopatham2016normalising,cometa,tutubalinafair}. However, after fine-tuning on the social media datasets (using the MS loss introduced earlier), \textsc{SapBert} outperforms SOTA significantly, indicating that knowledge acquired during the self-aligning pretraining can be adapted to a shifted domain without much effort.

\stitle{The \textsc{Adapter} Variant.} 
As an option for parameter efficient pretraining, we explore a variant of \textsc{SapBert} using a recently introduced training module named \textsc{Adapter}~ \citep{pmlr-v97-houlsby19a}. 
%
While maintaining the same pretraining scheme with the same \textsc{SapBert} online mining + MS loss, instead of training from the full model of \textsc{PubMedBert}, we insert new \textsc{Adapter} layers between Transformer layers of the fixed \textsc{PubMedBert}, and only train the weights of these \textsc{Adapter} layers. In our experiments, we use the enhanced \textsc{Adapter} configuration by \citet{pfeiffer-etal-2020-mad}. We include two variants 
where trained parameters are 13.22\% and 1.09\% of the full \textsc{SapBert} variant. The \textsc{Adapter} variant of \textsc{SapBert} achieves comparable performance to full-model-tuning in scientific datasets but lags behind in social media datasets, \Cref{tab:sap_bert_vs_sota}. The results indicate that 
more parameters are needed in pretraining for knowledge transfer to a shifted domain, in our case, the social media datasets.

\stitle{The Impact of Online Mining (\Cref{eq:mining}).}\label{sec:mining_exp}
As suggested in \Cref{Table:w_or_wo_mining}, switching off the online hard pairs mining procedure causes a large performance drop in $@1$ and a smaller but still significant drop in $@5$. This is due to the presence of many easy and already well-separated samples in the mini-batches. These uninformative training examples dominated the gradients and harmed the learning process. 

\begin{table}[H] 
\small
\vspace{-0.4em}
\centering
\begin{tabular}{lll}
\toprule
configuration & $@1$ & $@5$ \\
\midrule
Mining switched-on & \textbf{67.2} & \textbf{80.3} \\
Mining switched-off & 52.3$_{\downarrow 14.9}$ & 76.1$_{\downarrow 4.2}$ \\
\bottomrule
\end{tabular}
\vspace{-0.4em}
\caption{This table compares \textsc{PubMedBert+SapBert}'s performance with and without online hard mining on COMETA (zeroshot general). }
\label{Table:w_or_wo_mining}
\vspace{-0.8em}
\end{table}

\stitle{Integrating \textsc{SapBert} in Existing Systems.}
\textsc{SapBert} can be easily inserted into existing \textsc{Bert}-based MEL systems by initialising the systems with \textsc{SapBert} pretrained weights. We use the SOTA scientific language system, \textsc{BioSyn} (originally initialised with \textsc{BioBert} weights), as an example and show
the performance is boosted across all datasets (last two rows,  \Cref{tab:sap_bert_vs_sota}). 

\section{Conclusion}
We present \textsc{SapBert}, a 
self-alignment pretraining scheme for learning biomedical entity representations.
We highlight the consistent performance boost achieved by \textsc{SapBert}, obtaining new SOTA 
in all six widely used MEL benchmarking datasets. Strikingly, without any fine-tuning on task-specific labelled data, \textsc{SapBert} already outperforms the previous supervised SOTA (sophisticated hybrid entity linking systems) on multiple  datasets in the scientific language domain.  
%
Our work opens new avenues to explore for general domain self-alignment (e.g. by leveraging knowledge graphs such as DBpedia). We plan to incorporate other types of relations (i.e., hypernymy and hyponymy) and extend our model to sentence-level representation learning. In particular, our ongoing work using a combination of \textsc{SapBert} and \textsc{Adapter} is a promising direction for tackling sentence-level tasks.
\section*{Acknowledgements}
We thank the three reviewers and the Area Chair for their insightful comments and suggestions. FL is supported by Grace \& Thomas C.H. Chan Cambridge Scholarship. NC and MB would like to acknowledge funding from Health Data Research UK as part of the National Text Analytics project.

\bibliography{naacl2021}
\bibliographystyle{acl_natbib}

\appendix

\section{Evaluation Datasets Details}
\label{sec:datasets}

We divide our experimental datasets into two categories (1) scientific language datasests where the data is extracted from scientific papers and (2) social media language datasets where the data is coming from social media forums like \texttt{Reddit.com}. For an overview of the key statistics, see \Cref{Table:dataset_stats}.

\begin{table*} 
\small
\setlength{\tabcolsep}{3pt}
\centering
\begin{tabular}{lrrrrrrrrr}
\toprule
dataset &  NCBI & BC5CDR-d & BC5CDR-c & MedMentions & AskAPAtient & COMETA (s.g.) & COMETA (z.g.) \\
\midrule
Ontology & MEDIC & MEDIC & CTD & UMLS 2017AA & SNOMED \& AMT & SNOMED & SNOMED \\
$\mathcal{C}_{\text{searched}} \subsetneq \mathcal{C}_{\text{ontology}}  $? & \xmark & \xmark & \xmark & \xmark & \cmark & \xmark & \xmark \\
$|\mathcal{C}_{\text{searched}}|$ & 11,915 & 11,915 & 171,203 & 3,415,665 & 1,036 & 350,830 & 350,830\\
$|\mathcal{S}_{\text{searched}}|$ & 71,923 & 71,923 & 407,247 & 14,815,318 & 1,036 & 910,823 & 910,823 \\ 
$|\mathcal{M}_{\text{train}}|$ & 5,134 & 4,182 & 5,203 & 282,091 & 15,665.2 & 13,489 & 14,062\\
$|\mathcal{M}_{\text{validation}}|$ & 787 & 4,244 & 5,347 & 71,062 & 792.6 & 2,176 & 1,958\\ 
$|\mathcal{M}_{\text{test}}|$ & 960 & 4,424 & 5,385 & 70,405 & 866.2 & 4,350 & 3,995 \\ 
\bottomrule
\end{tabular}
\caption{This table contains basic statistics of the MEL datasets used in the study. $\mathcal{C}$ denotes the set of concepts; $\mathcal{S}$ denotes the set of all surface forms / synonyms of all concepts in $\mathcal{C}$; $\mathcal{M}$ denotes the set of mentions / queries. COMETA (s.g.) and (z.g.) are the stratified (general) and zeroshot (general) split respectively.}
\label{Table:dataset_stats}
\end{table*}

\subsection{Scientific Language Datasets}

\paragraph{NCBI disease \citep{dougan2014ncbi}} is a corpus containing 793 fully annotated PubMed abstracts and 6,881 mentions. The mentions are mapped into the MEDIC dictionary \citep{davis2012medic}. We denote this dataset as ``NCBI'' in our experiments.

\paragraph{BC5CDR \citep{li2016biocreative}} consists of 1,500 PubMed articles with 4,409 annotated chemicals, 5,818 diseases and 3,116 chemical-disease interactions. The disease mentions are mapped into the MEDIC dictionary like the NCBI disease corpus. The chemical mentions are mapped into the Comparative Toxicogenomics Database (CTD) \citep{davis2019comparative} chemical dictionary. We denote the disease and chemical mention sets as ``BC5CDR-d'' and ``BC5CDR-c'' respectively. For NCBI and BC5CDR we use the same data and evaluation protocol by \citet{sung2020biomedical}.\footnote{\url{https://github.com/dmis-lab/BioSyn}}

\paragraph{MedMentions \citep{mohan2018medmentions}} is a very-large-scale entity linking dataset containing over 4,000 abstracts and over 350,000 mentions linked to UMLS 2017AA. According to \citet{mohan2018medmentions}, training \textsc{TaggerOne} \citep{leaman2016taggerone}, a very popular MEL system, on a subset of MedMentions require >900 GB of RAM. Its massive number of mentions and more importantly the used reference ontology (UMLS 2017AA has 3M+ concepts) make the application of most MEL systems infeasible. However, through our metric learning formulation, \textsc{SapBert} can be applied
on MedMentions with minimal effort.

\subsection{Social-Media Language Datasets}

\paragraph{AskAPatient \citep{limsopatham2016normalising}} includes 17,324  adverse  drug  reaction  (ADR)  annotations collected from \texttt{askapatient.com} blog posts.  The mentions are  mapped  to  1,036  medical  concepts grounded onto SNOMED-CT \citep{donnelly2006snomed} and AMT (the Australian  Medicines Terminology). For this dataset, we follow the 10-fold evaluation protocol stated in the original paper.\footnote{\url{https://zenodo.org/record/55013}}

\paragraph{COMETA \citep{cometa}} is a recently released large-scale MEL dataset that specifically focuses on MEL in the social media domain, containing around 20k medical mentions extracted from health-related discussions on \texttt{reddit.com}. Mentions are mapped to SNOMED-CT. We use the ``stratified (general)'' split and follow the evaluation protocol of the original paper.\footnote{\url{https://www.siphs.org/corpus}}

\section{Model \& Training Details}
\label{sec:appendix_model_and_train}

\begin{table*}[t]
\setlength{\tabcolsep}{1.5pt}
    \centering
\small
\centering
\begin{tabular}{lccccccccccccccccccc}
\toprule
 \multirow{2}{*}{model} & \multicolumn{2}{c}{NCBI} &  &\multicolumn{2}{c}{BC5CDR-d} &  &\multicolumn{2}{c}{BC5CDR-c} & & \multicolumn{2}{c}{MedMentions} & & \multicolumn{2}{c}{AskAPatient} & 
& \multicolumn{2}{c}{COMETA}\\
\cmidrule[1.5pt]{2-3}\cmidrule[1.5pt]{5-6}\cmidrule[1.5pt]{8-9}\cmidrule[1.5pt]{11-12}\cmidrule[1.5pt]{14-15}\cmidrule[1.5pt]{17-18}
 & $@1$ & $@5$ & & $@1$ & $@5$ & & $@1$ & $@5$ & & $@1$ & $@5$ & & $@1$ & $@5$ & & $@1$ & $@5$  \\
\hline
 \textsc{Sieve-Based} \scriptsize \citep{d2015sieve} &  84.7 & - & & 84.1 & - & & 90.7 & - & & -&-\\
  \textsc{WordCNN} \scriptsize \cite{limsopatham2016normalising} & - & - & & - & - & & - & - & & - & - & & 81.4 & - & & - & - \\
 \textsc{WordGRU+TF-IDF} \scriptsize  \citep{tutubalina2018medical} & - & - & & - & - & & - & - & & - & - & &  85.7 & - & & - & -   \\
 \textsc{TaggerOne} \scriptsize \citep{leaman2016taggerone} &  87.7 & - & & 88.9 & - & & 94.1 & - &  & OOM & OOM & & - & -  & & -  & -\\
 \textsc{NormCo} \scriptsize \citep{wright2019normco} & 87.8 & - & & 88.0 & - &  & -&-&  & -&- & & - & -  & & -  & -\\
 \textsc{BNE} \scriptsize  \citep{phan2019robust} & 87.7 & -  & & 90.6 & - & & 95.8 & - &  & -&- & & - & -  & & -  & -\\
 \textsc{BertRank} \scriptsize \citep{ji2020bert} & 89.1 & - &  & -&-&  & -&-& & -&-  & & - & -  & & -  & - \\
  \textsc{Gen-Rank} \scriptsize \citep{xu-etal-2020-generate} & - & - & & - & - & & - & - & & - & - & & \textbf{87.5} & -  & & -  & - \\
 \textsc{BioSyn} \scriptsize \citep{sung2020biomedical} & \textbf{91.1} & \textbf{93.9} & & \textbf{93.2} & \textbf{96.0} & & \textbf{96.6} & \textbf{97.2} & & OOM & OOM & & 82.6$^\ast$ & 87.0$^\ast$ & & 71.3$^\ast$ & 77.8$^\ast$ \\ 
\textsc{Dict+Soilos+Neural} \scriptsize \citep{cometa} & - & - & & - & - & & - & - & & - & - & & - & -  & & \textbf{79.0}  & - \\
\hline
supervised SOTA & 91.1 & 93.9 & & 93.2 & 96.0 & & 96.6 & 97.2 & & OOM & OOM & & 87.5 & - & & 79.0 & - \\
\bottomrule
\end{tabular}
\caption{\small A list of baselines on the 6 different MEL datasets, including both scientific and social media language ones. The last row collects reported numbers from the best performing models. ``$\ast$'' denotes results produced using official released code. ``-'' denotes results not reported in the cited paper. ``OOM'' means out-of-memoery.}
\label{tab:sap_bert_vs_sota_full}
\end{table*}

\subsection{The Choice of Base Models} 
We list all the versions of \textsc{Bert} models used in this study, linking to the specific versions in \Cref{Table:model_url}. Note that we exhaustively tried all official variants of the selected models and the best performing ones are chosen. All \textsc{Bert} models refer to the \textsc{Bert}$_{\text{Base}}$ architecture in this paper.

\begin{table*}[t] 
\small
\setlength{\tabcolsep}{1pt}
\centering
\begin{tabular}{ll}
\toprule
model & URL \\
\midrule
vanilla \textsc{Bert} \citep{devlin2019bert} & \tiny \url{https://huggingface.co/bert-base-uncased} \\
\textsc{BioBert} \citep{lee2020biobert} &  \tiny\url{https://huggingface.co/dmis-lab/biobert-v1.1}\\
\textsc{BlueBert} \citep{peng2019transfer} &  \tiny\url{https://huggingface.co/bionlp/bluebert_pubmed_mimic_uncased_L-12_H-768_A-12} \\
\textsc{ClinicalBert} \citep{clinicalbert} &  \tiny\url{https://huggingface.co/emilyalsentzer/Bio_ClinicalBERT}\\
\textsc{SciBert} \citep{Beltagy2019SciBERT} &  \tiny\url{https://huggingface.co/allenai/scibert_scivocab_uncased}\\
\textsc{UmlsBert} \citep{umlsbert} &  \tiny\url{https://www.dropbox.com/s/qaoq5gfen69xdcc/umlsbert.tar.xz?dl=0} \\
\textsc{PubMedBert} \citep{pubmedbert} &  \tiny\url{https://huggingface.co/microsoft/BiomedNLP-PubMedBERT-base-uncased-abstract-fulltext} \\
\bottomrule
\end{tabular}
\caption{This table lists the URL of models used in this study. }
\label{Table:model_url}
\end{table*}

\subsection{Comparing Loss Functions}\label{sec:compare_loss}
We use COMETA (zeroshot general) as a benchmark for selecting learning objectives. Note that this split of COMETA is different from the stratified-general split used in \Cref{tab:sap_bert_vs_sota_full}. It is very challenging (so easy to see the difference of the performance) and also does not directly affect the model's performance on other datasets. The results are listed in \Cref{Table:compare_loss}. Note that online mining is switched on for all models here.

\begin{table}[H] 
\small
\setlength{\tabcolsep}{1pt}
\centering
\begin{tabular}{lcc}
\toprule
loss & $@1$ & $@5$ \\
\midrule
cosine loss \citep{phan2019robust} & 55.1 & 64.6 \\
max-margin triplet loss \citep{cometa} & 64.6 & 74.6 \\
NCA loss \citep{goldberger2005neighbourhood} & 65.2 & 77.0 \\
Lifted-Structure loss \citep{oh2016deep} & 62.0 & 72.1 \\
InfoNCE \citep{oord2018representation,he2020momentum} & 63.3 & 74.2 \\
Circle loss \citep{sun2020circle} & 66.7 & 78.7\\
\midrule
Multi-Similarity loss \citep{wang2019multi} & \textbf{67.2} & \textbf{80.3} \\
\bottomrule
\end{tabular}
\caption{This table compares loss functions used for \textsc{SapBert} pretraining. Numbers reported are on COMETA (zeroshot general).}
\label{Table:compare_loss}
\end{table}

The cosine loss was used by \citet{phan2019robust} for learning UMLS synonyms for LSTM models. The max-margin triplet loss was used by \citet{cometa} for training MEL models. A very similar (though not identical) hinge-loss was used by \citet{schumacher2020clinical} for clinical concept linking. InfoNCE has been very popular in self-supervised learning and contrastive learning \citep{oord2018representation,he2020momentum}. Lifted-Structure loss \citep{oh2016deep} and NCA loss \citep{goldberger2005neighbourhood} are two very classic metric learning objectives. Multi-Similarity loss \citep{wang2019multi} and Circle loss \citep{sun2020circle} are two recently proposed metric learning objectives and have been considered as SOTA on large-scale visual recognition benchmarks. 

\subsection{Details of \textsc{Adapter}s} 
In \Cref{Table:more_on_adapters} we list number of parameters trained in the three \textsc{Adapter} variants along with full-model-tuning for easy comparison. 

\begin{table}[H] 
\small
\setlength{\tabcolsep}{3pt}
\centering
\begin{tabular}{lccc}
\toprule
method & \scriptsize reduction rate & \#params & $\frac{\text{\#params}}{\text{\#params in }\textsc{Bert}}$  \\
\midrule
\textsc{Adapter$_{13\%}$} & 1 & 14.47M & 13.22\% \\
\textsc{Adapter$_{1\%}$} & 16 & 0.60M & 1.09\% \\
\midrule
full-model-tuning & - & 109.48M & 100\% \\
\bottomrule
\end{tabular}
\caption{This table compares number of parameters trained in \textsc{Adapter} variants and also full-model-tuning.}
\label{Table:more_on_adapters}
\end{table}

\subsection{Hardware Configurations}\label{sec:appendix_hardware}
All our experiments are conducted on a server with specifications listed in \Cref{Table:hardware}.
\begin{table}[H] 
\small
\setlength{\tabcolsep}{1pt}
\centering
\begin{tabular}{lr}
\toprule
hardware & specification \\
\midrule
 RAM & 192 GB \\
 CPU &  Intel Xeon W-2255 @3.70GHz, 10-core 20-threads\\
 GPU & NVIDIA GeForce RTX 2080 Ti (11 GB) $\times$ 4\\
\bottomrule
\end{tabular}
\caption{Hardware specifications of the used machine.}
\label{Table:hardware}
\end{table}

\begin{table*}[h] 
\small
\setlength{\tabcolsep}{1pt}
\centering
\begin{tabular}{lr}
\toprule
hyper-parameters & search space \\
\midrule
learning rate for pretraining \& fine-tuning \textsc{SapBert} & \{\texttt{1e-4}, \texttt{2e-5}$^\ast$, \texttt{5e-5}, \texttt{1e-5}, \texttt{1e-6}\} \\
pretraining batch size & \{128, 256, 512$^\ast$, 1024\} \\
pretraining training iterations & \{10k, 20k, 30k, 40k, 50k (1 epoch)$^\ast$, 100k (2 epochs)\} \\
fine-tuning epochs on scientific language datasets & \{1, 2, 3$^\ast$, 5\} \\
fine-training epochs on AskAPatient & \{5, 10, 15$^\ast$, 20\} \\
fine-training epochs on COMETA & \{5, 10$^\ast$, 15, 20\} \\
\texttt{max\_seq\_length} of \textsc{Bert} tokenizer & \{15, 20, 25$^\ast$, 30\} \\
$\lambda$ in Online Mining & \{0.05, 0.1, 0.2$^\ast$, 0.3\} \\
$\alpha$ in MS loss & \{1, 2 \citep{wang2019multi}$^\ast$, 3\} \\
$\beta$ in MS loss & \{40, 50 \citep{wang2019multi}$^\ast$, 60\} \\
$\epsilon$ in MS loss & \{0.5$^\ast$, 1 \citep{wang2019multi}\} \\
$\alpha$ in max-margin triplet loss  & \{0.05, 0.1, 0.2 \citep{cometa}$^\ast$, 0.3\} \\
softmax scale in NCA loss  & \{1 \citep{goldberger2005neighbourhood}, 5, 10, 20$^\ast$, 30\} \\
$\alpha$ in Lifted-Structured loss  & \{0.5$^\ast$, 1 \citep{oh2016deep}\} \\
$\tau$ (temperature) in InfoNCE  & \{0.07 \citep{he2020momentum}$^\ast$, 0.5 \citep{oord2018representation}\} \\
$m$ in Circle loss  & \{0.25 \citep{sun2020circle}$^\ast$, 0.4 \citep{sun2020circle}\} \\
$\gamma$ in Circle loss  & \{80 \citep{sun2020circle}, 256 \citep{sun2020circle}$^\ast$\} \\
\bottomrule
\end{tabular}
\caption{This table lists the search space for hyper-parameters used. $\ast$ means the used ones for reporting results.}
\label{Table:search_space}
\end{table*}

\section{Other Details}
\label{sec:appendix3}

\subsection{The Full Table of Supervised Baseline Models}\label{sec:appendix_full_baselines}
The full table of supervised baseline models is provided in \Cref{tab:sap_bert_vs_sota_full}.

\subsection{Hyper-Parameters Search Scope}\label{sec:hyperparam}
\Cref{Table:search_space} lists hyper-parameter search space for obtaining the set of used numbers. Note that the chosen hyper-parameters yield the overall best performance but might be sub-optimal on any single dataset. Also, we balanced the memory limit and model performance.

\subsection{A High-Resolution Version of \Cref{fig:front}}
\begin{figure*}
    \includegraphics[width=\linewidth]{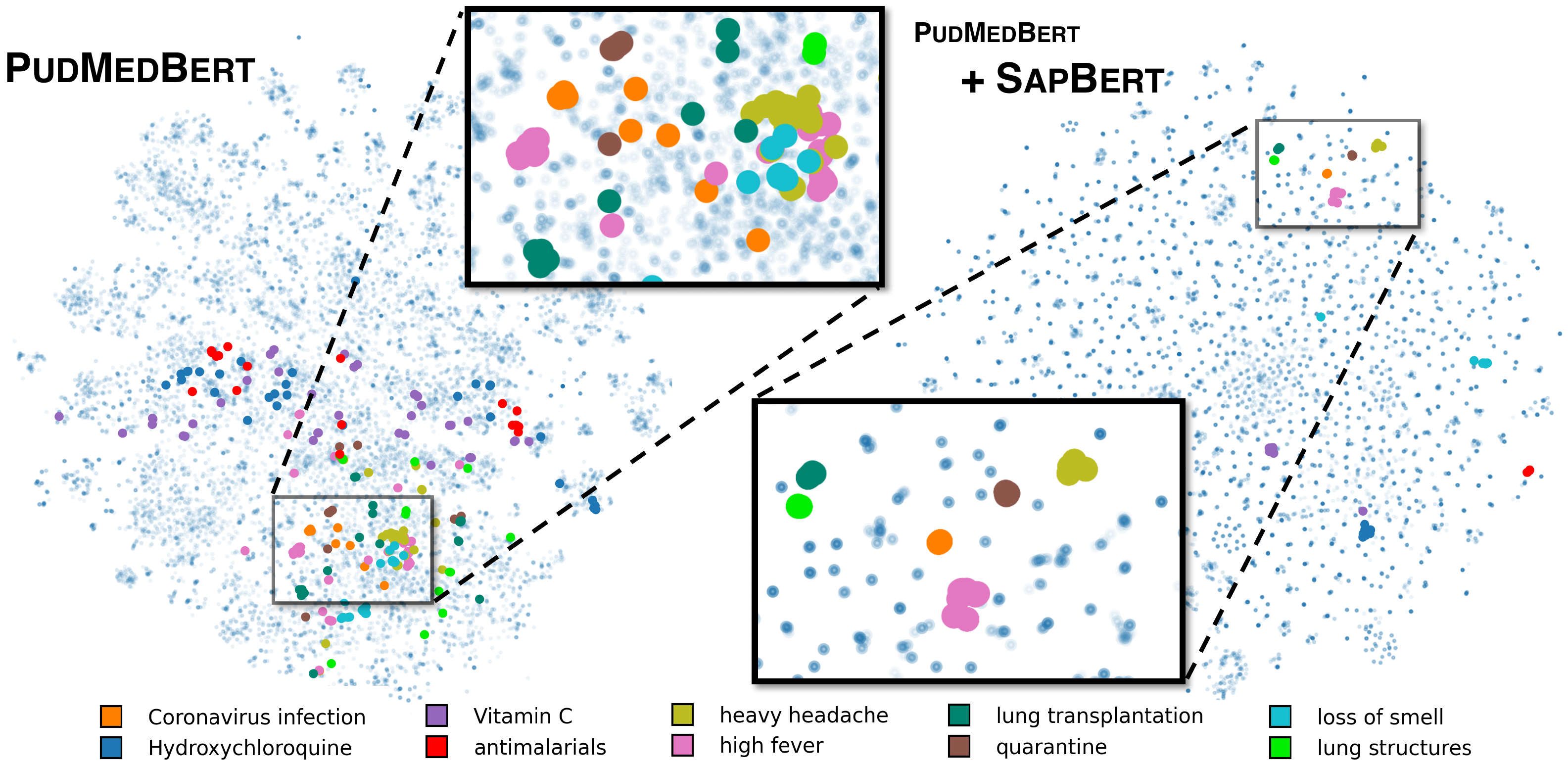}
    \caption{Same as \Cref{fig:front} in the main text, but generated with a higher resolution.}
    \label{fig:front_full}
\end{figure*}

We show a clearer version of t-SNE embedding visualisation in \Cref{fig:front_full}.




\end{document}